\documentclass[pdflatex,sn-basic]{sn-jnl}

\usepackage{graphicx}%
\usepackage{multirow}%
\usepackage{amsmath,amssymb,amsfonts}%
\usepackage{amsthm}%
\usepackage{mathrsfs}%
\usepackage[title]{appendix}%
\usepackage{xcolor}%
\usepackage{textcomp}%
\usepackage{manyfoot}%
\usepackage{booktabs}%
\usepackage{algorithm}%
\usepackage{algorithmicx}%
\usepackage{algpseudocode}%
\usepackage{listings}%
\usepackage{pifont}
\usepackage[english]{babel}
\usepackage[utf8x]{inputenc}
\usepackage[colorinlistoftodos]{todonotes}
\usepackage{subcaption}

\usepackage{array}      
\usepackage{tabularx}   
\usepackage{pdflscape}
\usepackage{makecell}

\theoremstyle{thmstyleone}%
\theoremstyle{thmstyletwo}%

\theoremstyle{thmstylethree}%

\makeatletter
\def\fnm#1{#1}
\def\sur#1{#1}
\def\orgdiv#1{#1}
\def\orgname#1{#1}
\def\orgaddress#1{#1}
\def\city#1{#1}
\def\postcode#1{#1}
\def\country#1{#1}
\makeatother

\begin{document}

\title[Decoupled Prototype Matching with Vision Foundation Models for Few-Shot Industrial Object Detection]{Decoupled Prototype Matching with Vision Foundation Models for Few-Shot Industrial Object Detection}

\author*[1]{\fnm{Hari Prasanth} \sur{S. M.}}\email{hariprasanth.sm@aalto.fi}

\author[1]{\fnm{Nilusha} \sur{Jayawickrama}}\email{nilusha.jayawickrama@aalto.fi}

\author[1]{\fnm{Risto} \sur{Ojala}}\email{risto.ojala@aalto.fi}

\affil[1]{\orgdiv{Department of Energy and Mechanical Engineering}, \orgname{Aalto University}, \orgaddress{
\city{Espoo}, 
\postcode{02150}, 
\country{Finland}}}


\abstract{ 
Industrial object detection systems typically rely on large annotated datasets, which are expensive to collect and challenging to maintain in industrial scenarios where the inventory of objects changes frequently.
This work addresses the challenge of few-shot object detection in such industrial scenarios, where only a limited number of labeled samples are available for newly introduced objects.
We present a detection framework that leverages vision foundation models to recognize objects with minimal supervision.
The method constructs class prototypes from a small set of reference samples by extracting feature representations.
For a given query scene during inference, object regions are generated using a segmentation model, and feature embeddings are extracted and matched with class prototypes using similarity matching.
We evaluate the detection method on three established industrial datasets from the Benchmark for 6D Object Pose Estimation benchmark following the official 2D object detection evaluation protocol.
We demonstrate competitive detection performance, improving AP by 6.9\% compared to the state-of-the-art training-free detection methods.
Furthermore, the presented method is able to onboard new objects using only a few reference images, without requiring any CAD models or large annotated datasets.
These properties make the approach well-suited for real-world industrial applications.
}

\keywords{Industrial Object Detection, Few-shot Detection, Vision Foundation Models, Prototype-based matching}



\maketitle

\bmhead{Acknowledgments}

This work was funded by Business Finland under the TwinFlow project (7374/31/2023). 
We gratefully acknowledge the computational resources provided by Aalto Science-IT.

\section{Introduction}\label{sec:introduction}


Visual perception has become a key technology enabler for industrial automation, robot manipulation, and inspections.
Object detection is a key perception task that supports a variety of subsequent applications such as bin-picking, quality inspection, grasp planning, and human-robot collaboration.
In comparison with generic object detection applications, industrial scenarios commonly consist of rigid objects with low texture, strong geometric similarities between categories, heavy occlusion and clutter. 

Although object detection has seen significant advancements via deep learning, most methods still rely on large-scale training data, with extensive bounding box or mask annotations for every object category.
Indeed, existing popular approaches have demonstrated state-of-the-art performance on general image benchmarks, but they fundamentally rely on abundant annotated data.
However, in real-world industrial applications, data annotation is costly, and often impractical due to frequent product changes and proprietary components.

Few-shot object detection (FSOD) has emerged as a solution in which new object classes are learned from a handful of labeled samples.
Many existing FSOD methods follow the base-novel class setting, where a detector is trained on base classes first with abundant labels and then fine-tuned on novel classes with a few support images.
Although these works have achieved impressive performance on benchmarks such as PASCAL VOC \citep{everingham2010pascal} and MS-COCO \citep{lin2014microsoft}, their assumptions may not transfer effectively to industrial scenarios.
Specifically, industrial objects often lack distinctive texture cues, with large number of classes similar to each other, and often appear in cluttered environment and heavy occlusion, making it challenging to differentiate objects using appearance cues alone.

In the context of industrial perception, previous work has identified the potential drawbacks of using an end-to-end detector-centric pipelines that couple object localization and classification~\citep{ren_faster_2017, qiao_defrcn_2021}.
This emphasizes the motivation for alternative frameworks that recognize objectness and learn representations at  the same time.
Recently, the vision foundation models that are trained with self-supervised or weakly-supervised learning on massive data have shown a strong ability to learn rich visual feature representations that capture semantic and geometric information about objects.
The most attractive properties of these models are their semantic consistency and transferability. 
These properties make them particularly promising for scenarios involving limited labeled data and unseen object categories.

Recent vision foundation models enable the construction of highly generalizable decoupled detection pipelines, where object localization and object identification are handled by separate modules~\citep{lu2025adapting}.
Rather than optimizing an object detector to jointly predict object regions and class labels, the object detection task is split into 
(i) class-agnostic proposal generation, which identifies potential object locations according to generic object cues, and 
(ii) object identification, which classifies these proposals according to robust object representations.
The decoupling formulation is well-suited for the few-shot scenarios, where it enables adding new classes without needing to update the localization module.

\begin{figure}[h]
\centering
\includegraphics[width=\linewidth]{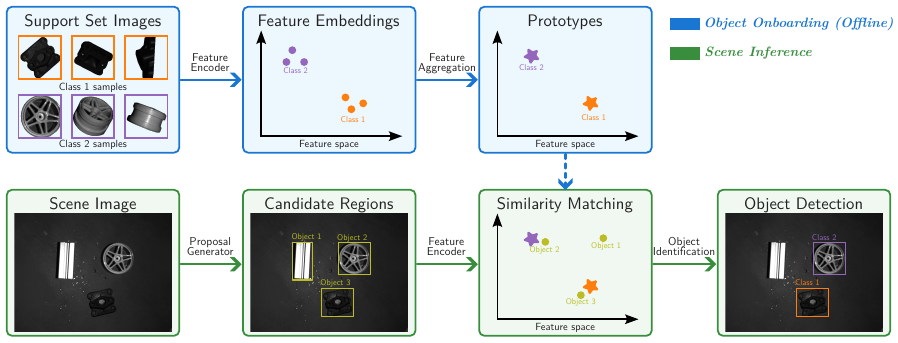}
\caption{Overview of the proposed DPM-VFM approach.}
\label{fig:overview}
\end{figure}

In this paper, we propose a Decoupled Prototype Matching approach using Vision Foundation Models (DPM-VFM) for few-shot object detection task for industrial environments.
An overview of the proposed DPM-VFM approach is presented in Figure~\ref{fig:overview}.
This approach leverages a segmentation-based vision foundation model to predict class agnostic object proposals from scene images and another vision foundation model to extract high quality visual embeddings of objects.
Given a few support samples per object class, class prototypes in the visual embedding space are constructed. During inference, object proposals are recognized by similarity-based matching against the constructed prototypes in the visual embedding space.
Our approach does not need retraining or fine-tuning for new object classes and is particularly suitable for industrial applications with rapidly evolving object classes.

The main contributions of this work are as follows:
\begin{itemize}
    \item We propose the DPM-VFM framework, a training-free, modular few-shot object detection framework designed for industrial environments that decouples object proposals and classifications using vision foundation models and prototype matching.
    \item We present a prototype-based object identification approach that leverages foundation-model feature embeddings for few-shot recognition with limited labeled data for fast onboarding.
    \item We provide experimental evaluation and ablation studies demonstrating the effectiveness of the DPM-VFM pipeline for data-efficient industrial object detection.
\end{itemize}

\section{Related Work}\label{sec:related_work}

\subsection{Object Detection}\label{subsec:object_detection}

Modern object detection based on fully supervised training has made tremendous development with deep learning.
Two-stage detectors~\citep{ren_faster_2017, he_mask_2017}, one-stage detectors~\citep{redmon_you_2016, lin_focal_2017}, and transformer-based architectures~\citep{carion_end--end_2020} have shown remarkable performance on large scale natural-image benchmarks where abundant labeled data is available.
However, the assumptions underlying these successes rarely apply to industrial environments, where objects tend to be texture-less, geometrically similar, and often appeared in cluttered and occluded scenes~\citep{drost_introducing_2017, kalra_towards_2024}. 
Moreover, collecting and annotating industrial data on a large-scale is expensive and time-consuming, especially when object inventories change frequently~\citep{hodan_bop_2018}.
These challenges motivate alternative approaches that can operate effectively with limited data.

FSOD aims to address the data scarcity problem by detecting novel object categories using only a few labeled samples.
The majority of existing methods follow a base-novel class paradigm, in which a detector is pretrained on a set of base classes, and subsequently finetuned to detect novel classes with a small support set \citep{yan_meta_2019, wang_frustratingly_2020}.
While this approach has demonstrated its effectiveness on natural-image benchmarks, there is still limited evidence of its applicability to industrial scenarios.

Existing FSOD approaches can be broadly divided into two categories: 
(i) Metric and Prototype-based approaches represent object classes as embeddings from support samples and perform classification by comparing similarities in an embedding space~\citep{snell_prototypical_2017, wu_meta-rcnn_2020}.
This approach allows new classes to be added without retraining the detector, but most methods have only been evaluated on VOC/COCO-style benchmarks.
(ii) Meta-learning and fine-tuning based approaches formulates FSOD as a task adaptation problem and use episodic training~\citep{yan_meta_2019, finn_model-agnostic_2017} or partial detector fine-tuning~\citep{wang_frustratingly_2020}.
More recent methods such as DeFRCN~\citep{qiao_defrcn_2021} and Meta-DETR~\citep{zhang_meta-detr_2021} further explore detector adaptation strategies for few-shot detection.
In practice, these methods require frequent retraining and are often sensitive to hyper-parameters, making them difficult to integrate into stable industrial detection pipelines.

\subsection{Segmentation-driven approaches} 

Object proposal generation plays an important role in detection pipelines as it is used to determine image regions to further processing. 
In detector-centric pipelines, proposals are learned together with the classification heads and are thus biased towards the object classes observed during training~\citep{ren_faster_2017}.
For few-shot and unseen object scenarios, this bias is a limitation, since the learned proposals do not generalize well to novel objects~\citep{qiao_defrcn_2021}.

To mitigate this, prior work has investigated class-agnostic objectness estimation, where object localization and identification are decoupled~\citep{o2015learning}.
Segmentation proposals offer a natural way to do this by learning to identify generic object boundaries rather than category-specific appearance cues~\citep{he_mask_2017}.
This is especially useful in cluttered scenes with densely packed or partially occluded objects.

Recent advances in vision foundation models have further strengthened segmentation-driven pipelines. 
The Segment Anything Model (SAM)~\citep{kirillov_segment_2023} provides strong class-agnostic object localization across a variety of domains, while self-supervised representation models such as DINO~\citep{caron_emerging_2021} learn transferable semantic embeddings that generalize well across tasks. 
These models offer complementary strengths: segmentation models predict where the objects are located, while representation models predict what the objects are, making it possible to develop segmentation-driven detection pipelines for few-shot recognition in industrial settings.

A baseline approach for segmentation-based industrial perception is CNOS~\citep{nguyen_cnos_2023}, 
which performs class-agnostic instance segmentation and relies on feature-matching using CAD-rendered templates. 
SAM-6D~\citep{lin_sam-6d_2024} similarly uses segmentation for localization followed by pose estimation. 
While these methods demonstrate strong performance, they primarily rely on CAD-based representations and do not directly address few-shot object detection from real image samples.
In addition, the use of rendered templates lacks the flexibility for those cases where no CAD models are available, or the immediate onboarding of novel objects from images is required.

\subsection{Research Gap}

Despite the recent advancements in vision foundation models, their applicability remains largely unexplored for few-shot industrial object detection.
In particular, their robustness in typical industrial scenes with low-texture objects, geometrically similar classes, and cluttered scenes has not been sufficiently studied.
Recent segmentation-driven pipelines, such as CNOS and SAM-6D, have achieved impressive detection performance on BOP industrial datasets.
However, their main focus is to recognize objects using the CAD-rendered templates. 
In many real-world applications, CAD models may not be always available, and new objects often need to be onboarded quickly. 
Therefore, it is essential to have a more flexible framework that enables few-shot object detection directly from real image samples.
In this work, we build on segmentation-driven detection approaches and propose a foundation model-based method that constructs prototype representations from a few support images to achieve few-shot object detection for industrial applications such as BOP benchmarks.

\section{Methods}\label{sec:methods}
\subsection{Problem Formulation}\label{subsec:problem_formulation}

We address the few-shot object detection problem with a decoupled localization and identification framework. 
A set of target object classes is defined as $\mathcal{C}=\{c_1, \dots, c_N\}$, where $N$ is the total number of classes. 
A support set for each class is defined as $\mathcal{S}_c = \{x_1^c, \dots, x_K^c\}$, where $x_i^c$ is the labeled image of the target object and $K$ is the number of labeled examples for each class. 
We retain only the object region in each support image $x_i^c \in \mathcal{S}_c$ by removing background pixels, so that each support sample represents only the target object instance.
There is no learning or finetuning of model parameters during onboarding of new classes. 
For a query scene image $I$ that contains one of more objects from $\mathcal{C}$, we aim to predict a set of detections $\mathcal{D}=(b_i, \hat{c}_i, s_i)_{i=1}^{M}$, where $b_i$ is a bounding box, $\hat{c}_i \in \mathcal{C}$ is the predicted class label, $s_i$ is a confidence score derived from embedding similarity matching, and $M$ is the total number of predicted detections. 
Unlike detector-centric few-shot object detection approaches that rely on base-class retraining and episodic fine-tuning \citep{yan_meta_2019, wu_meta-rcnn_2020}, our formulation assumes (i) no retraining or fine-tuning, (ii) no episodic optimization and (iii) no class-dependent proposal learning. Localization and identification are explicitly decoupled.

\subsection{Overview of the Decoupled Framework}\label{subsec:overview_decoupled_framework}

We approach object detection by factorizing the task into two independent components:
\begin{enumerate}
    \item Object proposal generation, which identifies object-like regions via a class-agnostic segmentation model.
    \item Similarity-based object identification, which assigns a class label to each selected proposal by finding the closest match in the prototype representation space.
\end{enumerate}

\begin{figure}[!h]
    \centering
    \includegraphics[width=\linewidth]{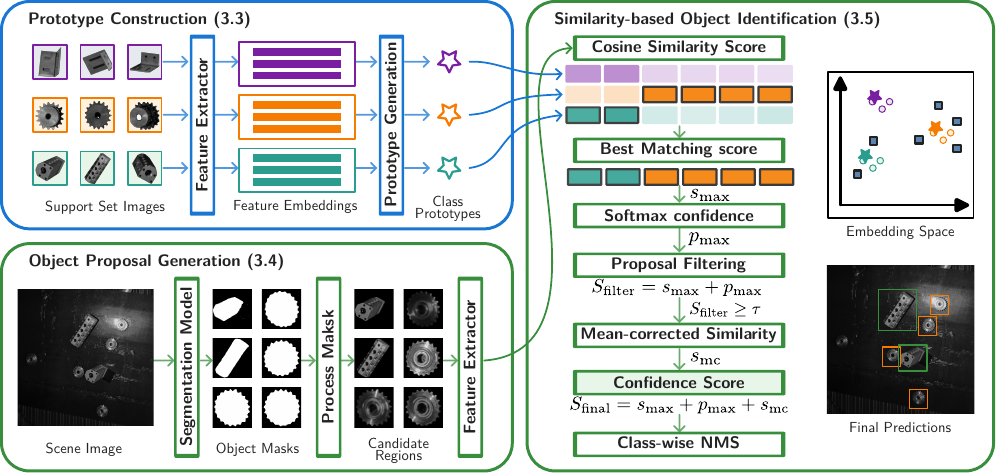}
    \caption{Architecture of the DPM-VFM pipeline.}
    \label{fig:pipeline_architecture}
\end{figure}

An overview of the proposed pipeline is illustrated in Figure~\ref{fig:pipeline_architecture}. The framework consists of (i) an offline prototype construction stage, where class prototypes are generated from support images (Section~\ref{subsec:prototype_construction}), and (ii) an online inference stage, where object proposals are extracted from the input scene (Sections~\ref{subsec:object_proposal_generation}) and subsequently identified via similarity-based matching (Sections~\ref{subsec:similarity_based_identification}).


\subsection{Prototype Construction}\label{subsec:prototype_construction}
For each class $c \in \mathcal{C}$, we construct the prototype from the images of the corresponding support set $\mathcal{S}_c$. 
For every image in the support set $x_i^c \in \mathcal{S}_c$, a semantic representation is obtained by using the DINO-based vision transformer~\citep{oquab_dinov2_2024, simeoni_dinov3_2025}.
Specifically, the feature embedding is obtained from the CLS token, which provides a global representation of the input image.
For each support set image $x_i^c$, the feature embedding is denoted as $z_i^c = f(x_i^c) \in \mathbb{R}^d$, where $f(.)$ represents the DINO feature extractor and $d$ is the embedding dimension.
Then, we perform $\ell_2$ normalization to compute the normalized feature descriptor $\Tilde{z}_i^c$.
The prototype for a class is computed by averaging normalized support embeddings: $p_c = \frac{1}{K} \sum_{i=1}^{K} \tilde{z}_i^c$.
This formulation is inspired by the idea of Prototypical Networks \citep{snell_prototypical_2017}, where a class is represented as a centroid in the embedding space.
Unlike meta-learning approaches, episodic training is not needed here.

\subsection{Object Proposal Generation}\label{subsec:object_proposal_generation}

We leverage SAM to generates class-agnostic object proposals. 
For a query scene image $I$, SAM generate a list of predicted masks $\mathcal{M}=\{m_1, \dots, m_P\}$ where each mask $m_i$ defines a potential object region. 
We use automatic mask generator from SAM to generate the object proposals. 
In the industrial applications, where objects tend to have low texture, high geometric similarity, and severe occlusion, segmentation-based objectness provides a natural way to identify candidate object regions based on generic object boundaries.
Unlike region proposal networks learned together with a classifier, segmentation-based proposals are free of class bias.

For each mask $m_i$, we compute the axis-aligned bounding box $b_i$ by extracting the minimal enclosing rectangle of all foreground pixels in the mask. 
In order to filter out noisy proposals, we (i) filter by minimum area, (ii) filter by confidence score, and (iii) apply non-maximum suppression (NMS) with threshold $\theta_{NMS}$.
Once the proposals are filtered, the corresponding object is extracted from the scene image. 
For each proposal bounding box $b_i$, a cropped region $x_i$ is extracted from the scene image, with background pixels are set to 0 to filter out the object away from foreground mask pixels, which could be another object or background.
The set of object proposals are denoted as $\mathcal{P} = \{x_1, \dots, x_{P'}\}$.

\subsection{Similarity-based Object Identification}\label{subsec:similarity_based_identification}

To obtain transferable semantic representations for object proposals, we use the same DINO model as for the class prototype generation.
For each object proposal $x_i$, the feature descriptor $\Tilde{z}_i$ is computed using DINO feature extraction and normalized. 
Then, we calculate cosine similarities for every proposal embeddings with all class prototypes:

\begin{equation}
    s_{i,c} = \tilde{z}_i^\top p_c
\end{equation}

The predicted class is found via nearest prototype  matching, which has the highest similarity score $\hat{c} = \arg \max_{c \in \mathcal{C}} s_{i, c}$,
and we denote the corresponding maximum similarity as $s_{\max} = s_{i,\hat{c}_i}$.

\subsubsection{Proposal Filtering}
We define a filtering score that combines similarity magnitude and class confidence. We compute a softmax over similarity scores:

\begin{equation} 
    p_{\max} = \frac{\exp(s_{i,\hat{c}_i})}{\sum_{c'} \exp(s_{i,c'})} 
\end{equation}

The filtering score is given by:
\begin{equation}
    S_{\mathrm{filter}} = s_{\max} + p_{\max}
\end{equation}

A proposal is retained only if the filtering score is greater than a filter threshold, i.e. $ S_{\mathrm{filter}} \ge \tau $.

\subsubsection{Confidence Estimation}
For retained proposals, we refine the confidence score by incorporating the relative separation of the top match. We define the mean-corrected similarity as:

\begin{equation} 
    s_{\mathrm{mc}} = s_{\max} - \frac{1}{|\mathcal{C}|} \sum_{c} s_{i,c} 
\end{equation}

The final confidence score is defined as:

\begin{equation} 
    S_{\mathrm{final}} = s_{\max} + p_{\max} + s_{\mathrm{mc}} 
\end{equation}

This formulation decouples proposal selection and confidence estimation, where filtering suppresses ambiguous detections and mean-corrected similarity improves ranking among retained proposals.
Finally, a class-wise NMS is applied using an IoU threshold to remove duplicate objects with high overlap.

\section{Experiments}\label{sec:experiments}

\subsection{Experimental Setup}\label{subsec:exp_setup}

We evaluate on three industrial datasets from the Benchmark for 6D Object Pose Estimation (BOP): ITODD~\citep{drost_introducing_2017}, IPD~\citep{kalra_towards_2024}, and XYZ-IBD~\citep{huang_xyz-ibd_2025}.
The BOP benchmark was originally introduced to establish standardized evaluation protocols for 6D pose estimation of rigid objects~\citep{hodan_bop_2018}. 
Since then, the benchmark has been evolved to support 2D object detection and instance segmentation tasks, and has expanded to include industrial datasets through BOP-Industrial track~\citep{hodan_bop_2024, nguyen_bop_2025}.

\begin{figure}[!h]
\centering

\begin{subfigure}[t]{0.32\linewidth}
    \centering
    \includegraphics[width=\linewidth]{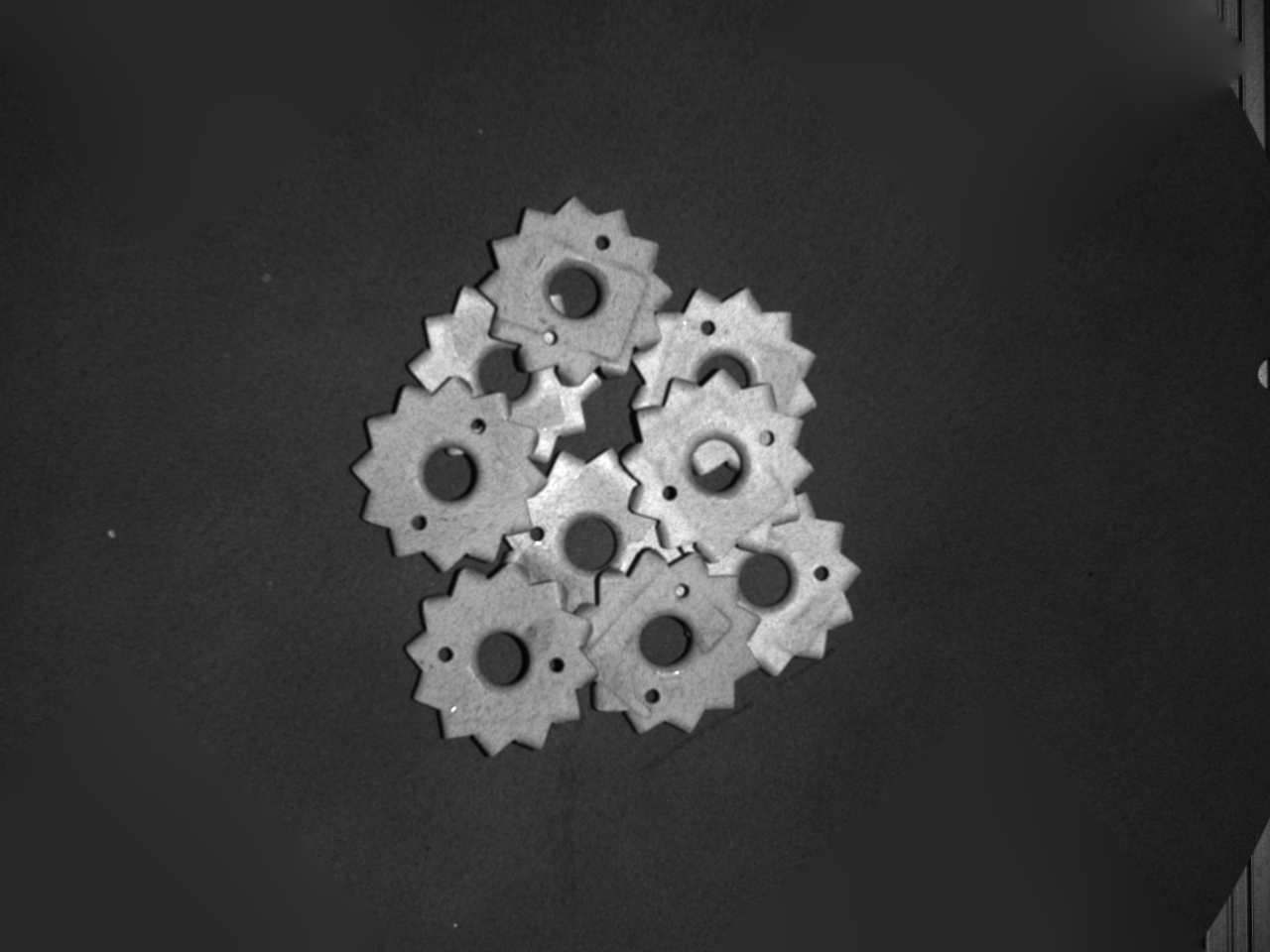}
    \caption{ITODD}
\end{subfigure}
\hfill
\begin{subfigure}[t]{0.32\linewidth}
    \centering
    \includegraphics[width=\linewidth]{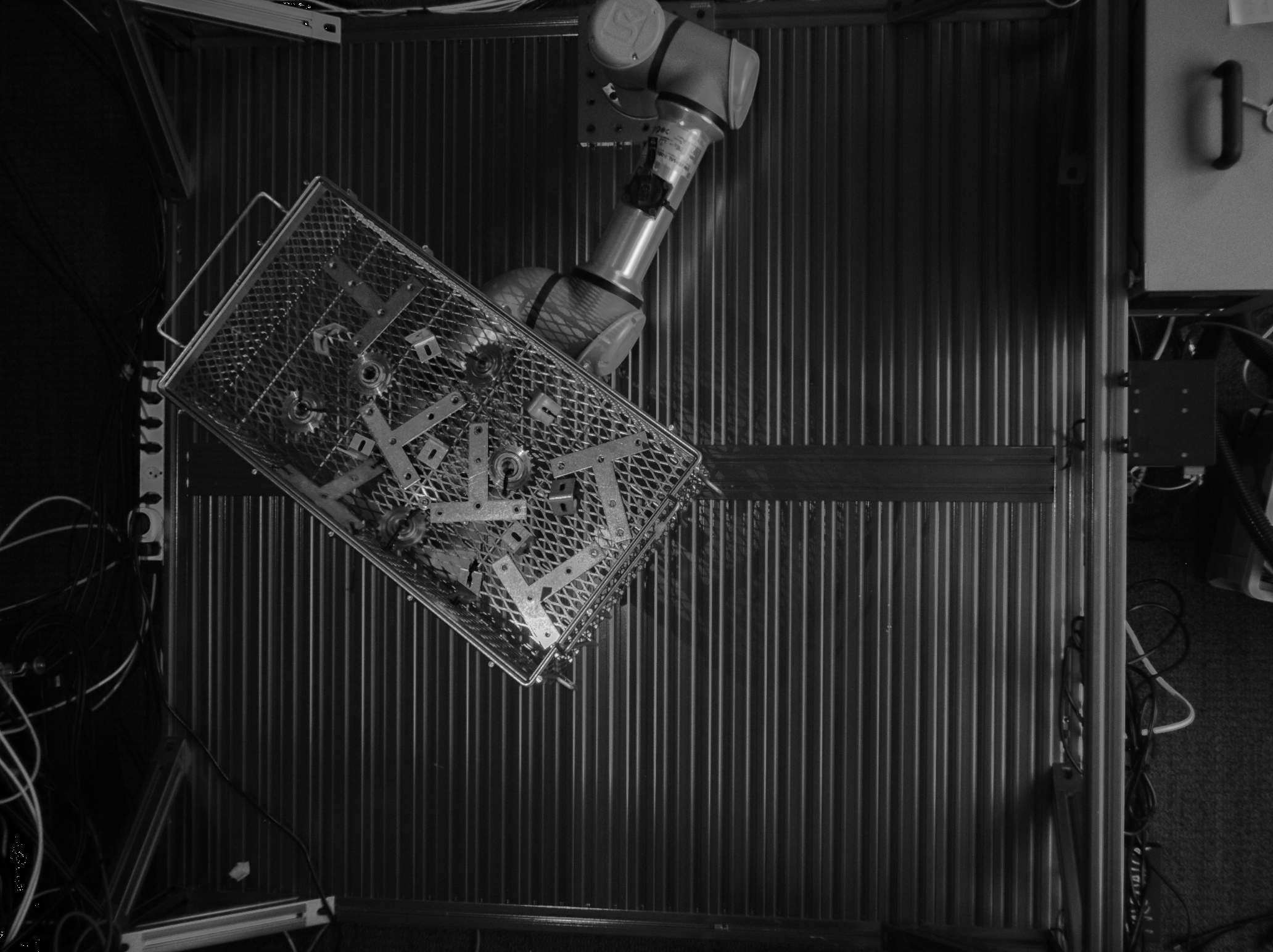}
    \caption{IPD}
\end{subfigure}
\hfill
\begin{subfigure}[t]{0.32\linewidth}
    \centering
    \includegraphics[trim={6cm 0cm 5cm 0cm}, clip, width=\linewidth]{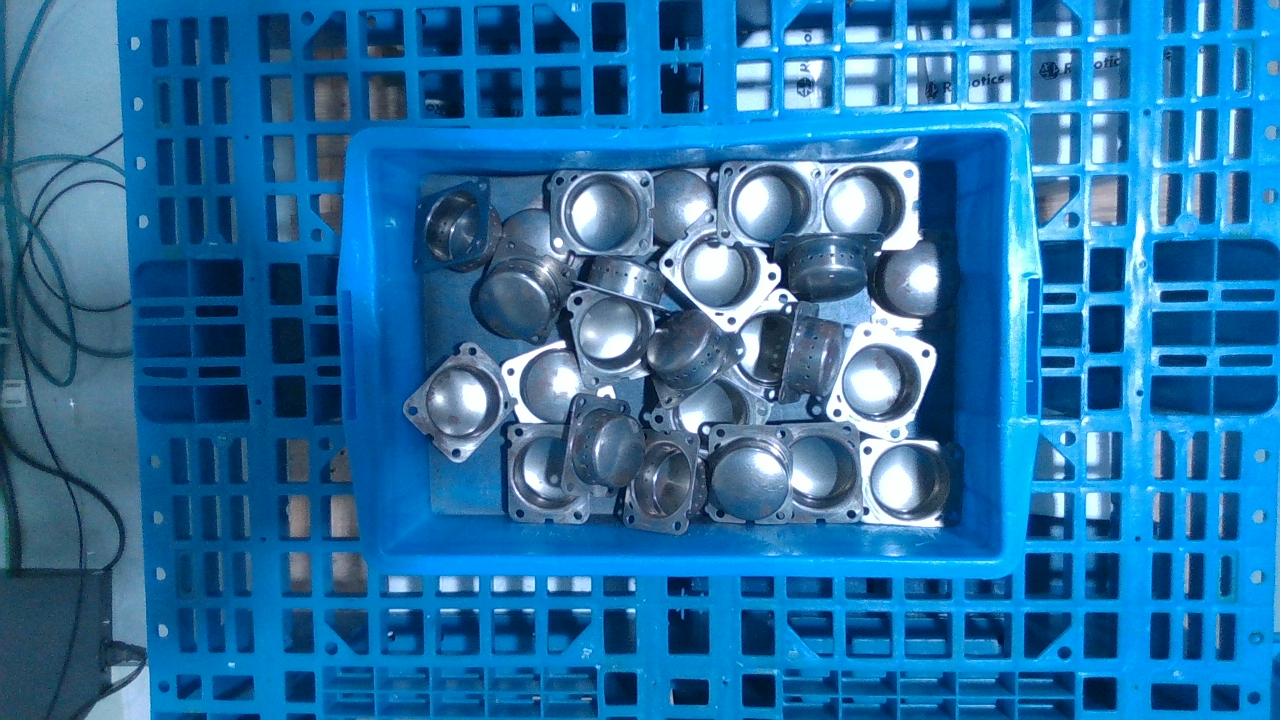}
    \caption{XYZ-IBD}
\end{subfigure}

\caption{Representative test images from the BOP industrial datasets used in this work.}

\label{fig:bop_dataset_examples}

\end{figure}

These datasets represent essential aspects of industrial perception tasks commonly found in manufacturing and bin-picking scenarios, such as dense clutters, reflective surfaces, noisy sensor data, and object classes with very similar appearance. 
This makes them very challenging for object detection techniques that need to generalize to unseen object classes with a limited labeled images. 
Basic dataset statistics are summarized in Table~\ref{tab:bop_datasets}, and representative test images from each dataset are presented in Figure~\ref{fig:bop_dataset_examples}.

\begin{table}[!h]
\centering
\caption{BOP-Industrial datasets used in evaluation.}
\label{tab:bop_datasets}
\begin{tabular}{l c c c l}
\toprule
Dataset & \# Objects & \# Scenes & \# Images & Key challenges \\
\midrule
ITODD   & 28 & 1 & 721 & Realistic bin-picking scenarios \\
IPD     & 10 & 15 & 1232 & Varying lighting conditions \\
XYZ-IBD & 15 & 60 & 297 & Texture-less objects with heavy occlusions \\
\bottomrule
\end{tabular}
\end{table}

In this work, we consider only the 2D object detection task and not the 6D pose estimation task.
All experiments are conducted using single-view RGB images, following common practice in recent industrial detection pipelines, to allow a fair comparison with existing approaches that have been evaluated on the same benchmarks.
The predicted detections are formatted according to the BOP detection protocol and submitted to the official evaluation server\footnote{\url{https://bop.felk.cvut.cz/home/}}, on which the final evaluation metrics are computed since the ground-truth annotations of the test sets are not publicly available.
We evaluate our method using the Average Precision (AP) metrics, following the BOP evaluation protocol. 
The AP metric is calculated as the mean of AP at different Intersection-over-Union (IoU) thresholds ranging from 0.50 to 0.95 with a step size of 0.05~\citep{sundermeyer2023bop}.

For each object class, ten sample instances are extracted by cropping the corresponding objects from the validation dataset. 
When the validation set does not contain enough instances, additional samples are obtained from the test set. 
The extracted images are resized while preserving the aspect ratio, where the longer edge is resized to 224 and the shorter edge is padded.
Segmentation masks for these sample images are applied using Roboflow’s annotation tool~\footnote{\url{https://roboflow.com/annotate}}. 
The support set sample images are created by applying the mask to the resized image, retaining only the object region while all background pixels are set to 0.

For proposal generation, we use SAMv1 (ViT-H), and for feature extraction, we use DIVOv3 (ViT-L/16).
The hyper-parameters of the segmentation model are adjusted to improve the quality of object proposals, while maintaining consistency across all datasets.
The feature embeddings are obtained using the CLS token from the DINO model outputs.
Table~\ref{tab:implementation_details} lists all the implementation details and hyper-parameters used in the DPM-VFM pipeline.

\begin{table}[t]
\caption{Implementation details and hyper-parameters of the DPM-VFM pipeline.}
\label{tab:implementation_details}
\centering
\small
\setlength{\tabcolsep}{6pt}
\renewcommand{\arraystretch}{1.12}
\begin{tabular}{ll}
\toprule
\textbf{Component} & \textbf{Configuration} \\
\midrule

\multicolumn{2}{l}{\textit{Support Set}} \\
$\#$ samples per class & 10 \\
Input resolution & $224 \times 224$ \\
\\

\multicolumn{2}{l}{\textit{Proposal Generators}} \\
SAMv1 & points: 16, $\theta_{\text{IoU}}$: 0.60, $\theta_{\text{stab}}$: 0.85 \\
SAMv2.1 & points: 16, $\theta_{\text{IoU}}$: 0.60, $\theta_{\text{stab}}$: 0.85 \\
FastSAM & $\theta_{\text{conf}}$: 0.10, $\theta_{\text{IoU}}$: 0.95 \\
Min. area ratio & 0.0005 \\
NMS threshold ($\theta_{\mathrm{NMS}}$) & 0.75 \\
\\

\multicolumn{2}{l}{\textit{Final Configuration}} \\
Segmentation model & \textbf{SAMv1 (ViT-H)} \\
Feature extractor & \textbf{DINOv3 (ViT-L/16)} \\
Filter threshold ($\tau$) & 0.4 \\
Post-NMS threshold & 0.5 \\

\bottomrule
\end{tabular}
\end{table}

\subsection{Performance Evaluation}\label{subsec:performance_evaluation}

We compare DPM-VFM framework against the state-of-the-art approaches CNOS~\citep{nguyen_cnos_2023} and SAM-6D~\citep{lin_sam-6d_2024}.
For a fair comparison, we use the same support samples and the same vision foundation models.
Also, we follow the official implementation settings and hyper-parameters reported in the respective works.
Results for the accuracy benchmark for all datasets are presented in Table~\ref{tab:bop_industrial_results}.
The proposed DPM-VFM method achieves competitive performance, outperforming existing approaches in most configurations.
In particular, DPM-VFM achieves higher AP on the IPD and XYZ-IBD datasets, demonstrating its effectiveness on textureless objects, varying lighting conditions and heavy occlusions. 

\begin{table}
\centering
\caption{\textbf{Performance comparison on BOP Industrial datasets.} 
The AP metric (higher is better) is reported here.
Each SOTA method is evaluated with two set of foundation model frameworks.
Bolded values show the best results.}
\label{tab:bop_industrial_results}
\begin{tabular}{llcccc}
\toprule
\multirow{2}{*}{Foundation Models} & 
\multirow{2}{*}{Method} &
\multicolumn{3}{c}{BOP Industrial} & 
\multirow{2}{*}{Mean} \\
\cmidrule(lr){3-5}
 &  & ITODD & IPD & XYZ-IBD &  \\
\midrule

\multirow{3}{*}{SAMv1 + DINOv2} 
    & CNOS~\citep{nguyen_cnos_2023} & 44.3 & 23.8 & 22.4 & 30.2 \\
    & SAM6D~\citep{lin_sam-6d_2024} & \textbf{48.1} & 33.8 & 31.3 & 37.7 \\
    & DPM-VFM                       & 45.2 & \textbf{49.5} & \textbf{37.1} & \textbf{43.9} \\

\midrule

\multirow{3}{*}{FastSAM + DINOv2} 
    & CNOS~\citep{nguyen_cnos_2023} & 48.3 & 31.5 & 27.9 & 35.9 \\
    & SAM6D~\citep{lin_sam-6d_2024} & \textbf{49.9} & 31.9 & 26.7 & 36.2 \\
    & DPM-VFM                       & 45.6 & \textbf{48.8} & \textbf{34.4} & \textbf{42.9} \\

\midrule
{SAMv1 + DINOv3} 
    & DPM-VFM                    & \textbf{46.0} & \textbf{51.5} & \textbf{36.4} & \textbf{44.6} \\

\bottomrule
\end{tabular}
\end{table}

\begin{figure}
    \centering
    \includegraphics[width=0.95\linewidth]{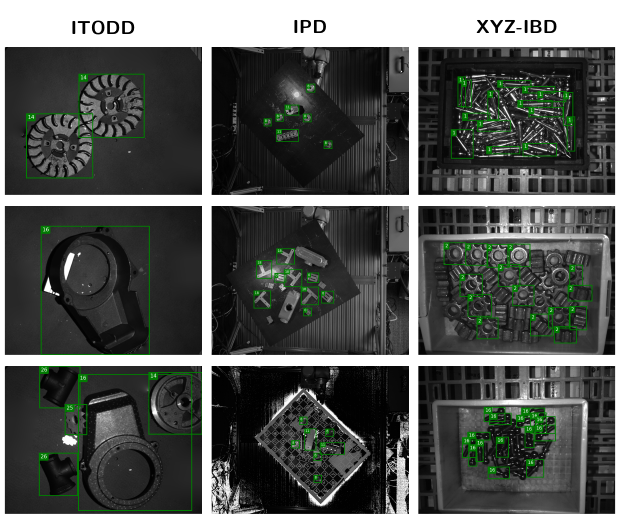}
    \caption{\textbf{Qualitative results of successful detections produced by DPM-VFM across BOP-industrial datasets.} Each row shows a different scene image, with three representative successful examples per dataset. Only the highest-confidence predictions are shown for visual clarity. Bounding boxes are color-coded as green (true positives), with class identifiers displayed at the top-left of each box.}
    \label{fig:qualitative_good_results}
\end{figure}

\begin{figure}
    \centering
    \includegraphics[width=0.95\linewidth]{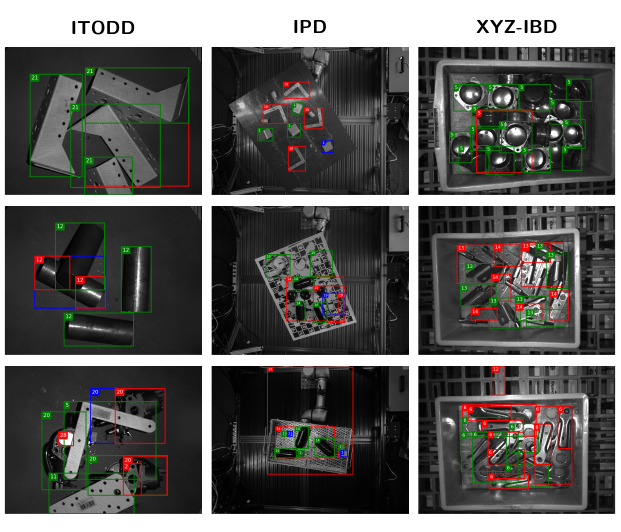}
    \caption{\textbf{Qualitative analysis of failure cases produced by DPM-VFM across BOP-industrial datasets.} Each row shows a different scene image, with three representative failure examples per dataset. Only the highest-confidence predictions are shown for visual clarity. Bounding boxes are color-coded as green (true positives), red (false positives), and blue (false negatives), with class identifiers displayed at the top-left of each box.}
    \label{fig:qualitative_bad_results}
\end{figure}

\begin{figure}
    \centering
    \includegraphics[width=\linewidth]{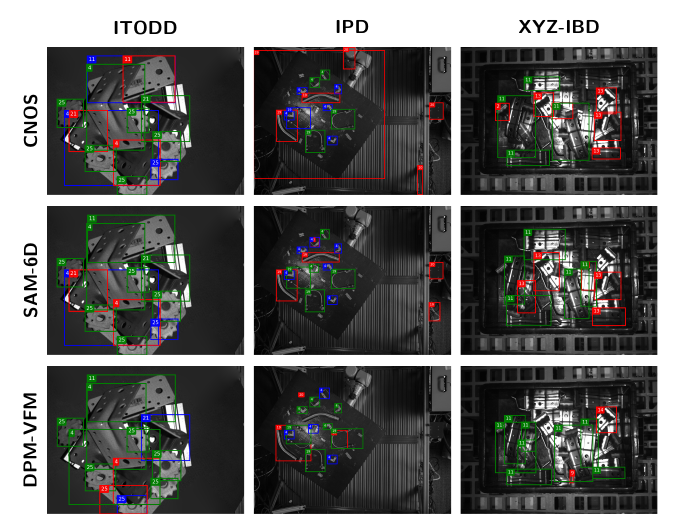}
    \caption{\textbf{Qualitative comparison with state-of-the-art methods on representative scene images.} Each row presents predictions from a different method: CNOS (top), SAM-6D (middle), and DPM-VFM (bottom). All methods are evaluated on the same scenes to enable direct comparison. Bounding boxes are color-coded as green (true positives), red (false positives), and blue (false negatives), with class identifiers displayed at the top-left of each box.}
    \label{fig:sota_comparison}
\end{figure}

Figure~\ref{fig:qualitative_good_results} presents few example successful detection cases of DPM-VFM predictions across all datasets, while Figure~\ref{fig:qualitative_bad_results} illustrates failure cases highlighting common challenges. 
For clarity, only the top-confidence predictions are visualized in each scene.
Bounding boxes are color-coded as green for true positives, red for false negatives, and blue for false negatives.
For the XYZ-IBD scenes, only the top 10 detections are displayed to improve visual readability.
In the IPD dataset, scenes often contains objects outside the target classes, which are not assigned class labels in the visualization. 
Additionally, Figure~\ref{fig:sota_comparison} presents qualitative comparisons with state-of-the-art methods on selected scene images.
The examples are selected from challenging cases based on qualitative inspection, where state-of-the-art methods exhibit lower detection performance.

\subsection{Prototype Representation Analysis}

\begin{figure}
    \centering
    \includegraphics[width=\linewidth]{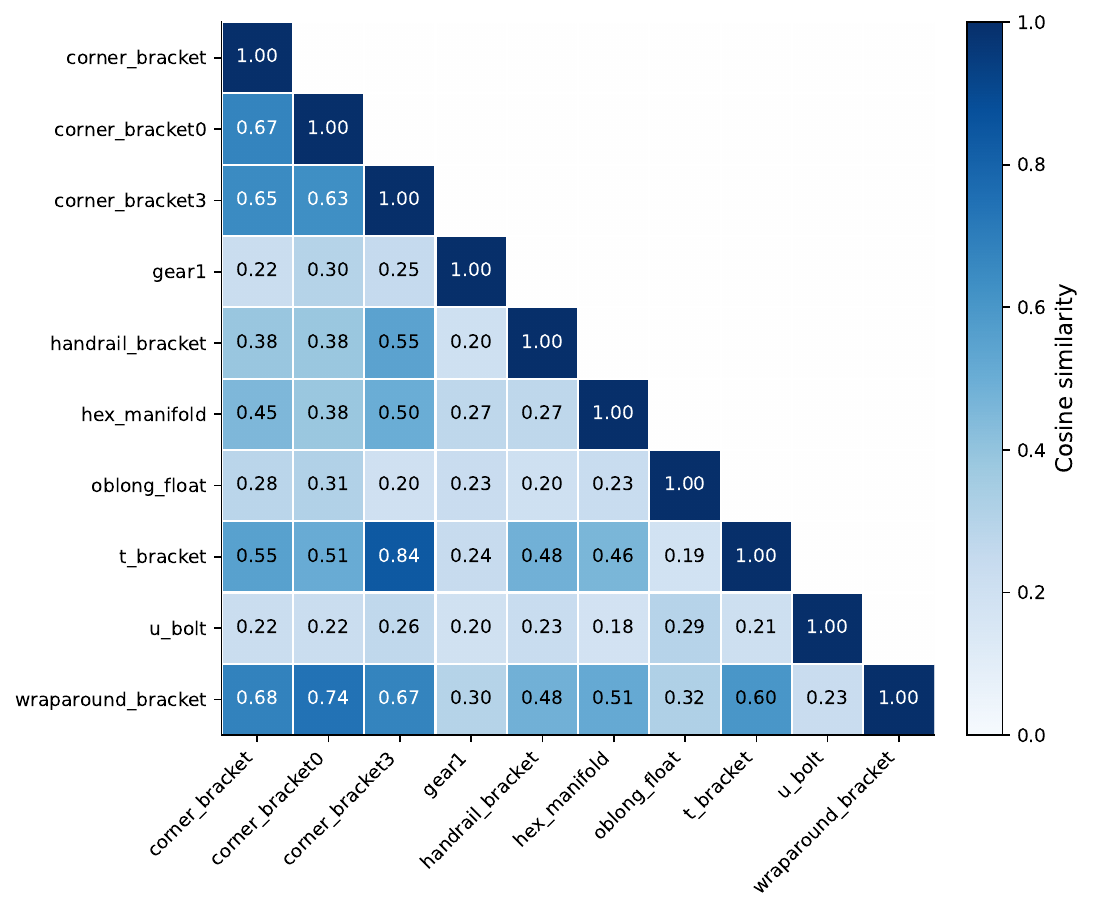}
    \caption{\textbf{Pairwise cosine similarity between class prototypes on the IPD dataset.} Each class is represented by its learned prototype. Only the lower triangular matrix is shown due to symmetry.}
    \label{fig:ipd_prototype_similarity}
\end{figure}

\begin{figure}
    \centering
    \includegraphics[width=\linewidth]{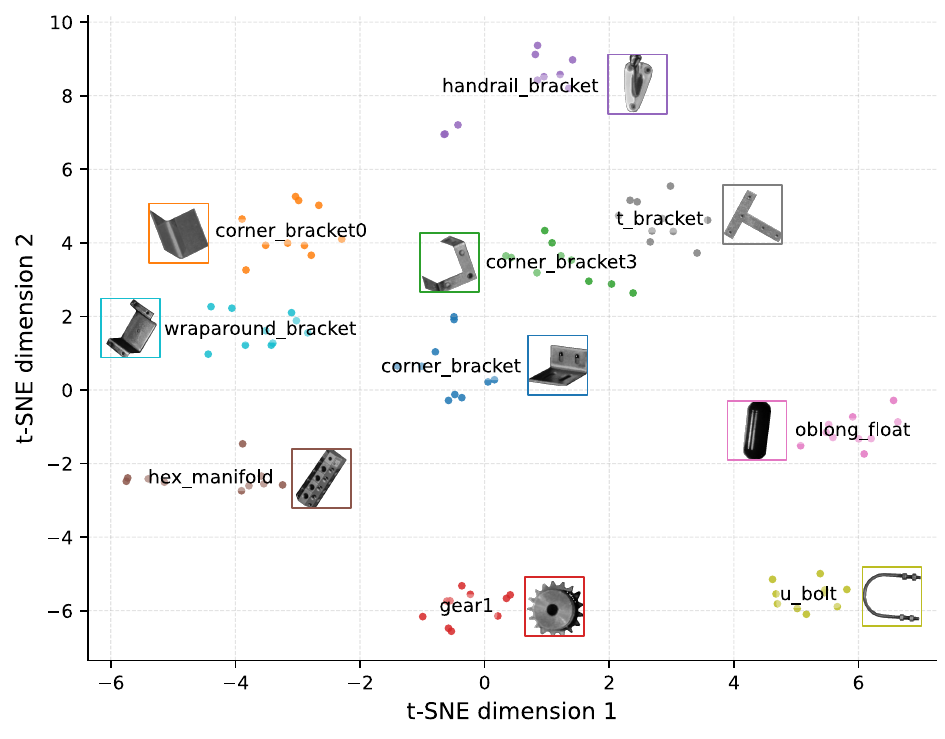}
    \caption{\textbf{t-SNE visualization of DINOv3 (ViT-L/16) CLS embeddings for the IPD dataset support set images.} Each point corresponds to one support set image (10 samples per class across 10 classes). Colors indicate object classes, labels are placed at the cluster centers, and the object is visualized next to the cluster. The embeddings exhibit clear class-wise clustering.}
    \label{fig:ipd_samples_embeddings}
\end{figure}

Figure~\ref{fig:ipd_prototype_similarity} illustrates the pairwise similarity between class prototypes in the IPD dataset, while Figure \ref{fig:ipd_samples_embeddings} illustrates the distribution of the DINOv3 feature embeddings of the IPD support set images in the feature space by reducing the dimensionality with t-SNE.
The embeddings of the same object class are color-coded for the visualization; however, the vision foundation model has no prior information about the class identification.
Each object class forms a compact cluster with their 10 support sample images in the embedding space, showing the discriminative property of the feature embeddings for corresponding object class. 

\subsection{Ablation Studies}
\label{subsec:ablation}

\subsubsection{Effect of vision foundation model variants}
\label{subsubsec:ablation_vf}

\begin{figure}
    \centering
    \includegraphics[width=\linewidth]{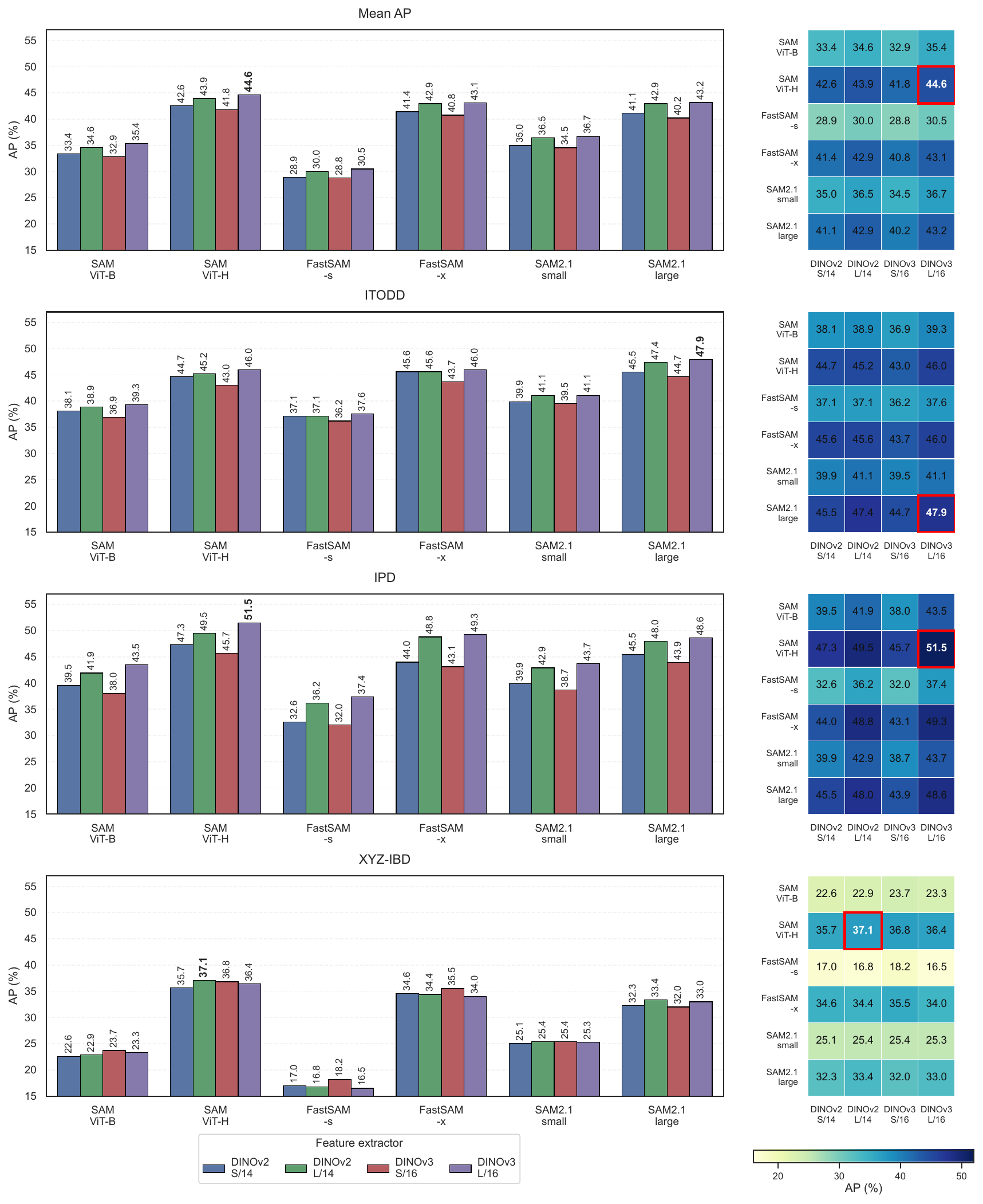}
    \caption{\textbf{Ablation study of segmentation models and feature extractors in the proposed DPM-VFM pipeline across BOP-industrial datasets.} Each row reports AP for a dataset (Mean, ITODD, IPD, XYZ-IBD), while columns correspond to different feature extractors. Left: grouped bar charts; right: corresponding heatmaps. The best-performing configuration in each dataset is highlighted in bold.}
    \label{fig:vfm_ablation}
\end{figure}

We evaluate the impact of different vision foundation model variants within the DPM-VFM pipeline, as illustrated in Fig.~\ref{fig:vfm_ablation}. 
Specifically, DINOv2~\citep{oquab_dinov2_2024} and DINOv3~\citep{simeoni_dinov3_2025} are used for feature extractions, while SAMv1~\citep{kirillov_segment_2023}, FastSAM~\citep{zhao_fast_2023}, and SAMv2.1~\citep{ravi_sam_2024} are used for class-agnostic proposal generation. 
For each family of models, smaller and larger variants are considered to analyze the effect of model size. 
This study provides insights into the segmentation quality and feature representation in determining final detection performance.

\subsubsection{Effect of Hyper-parameters}

We conduct ablation studies to validate the sensitivity of the DPM-VFM pipeline with SAMv1 and DINOv3 with respect to the filtering threshold $\tau$ and post-processing class-wise NMS threshold. 
The AP on the three benchmarks, ITODD, IPD and XYZ-IBD, as well as the mean AP across the datasets are reported. 
For each ablation study, only one of the parameters is changed, while the other is fixed. 
The results are summarized in Table~\ref{tab:thresh_ablation} and Table~\ref{tab:nms_ablation}.

\begin{table}
\caption{Ablation study of filtering threshold $\tau$. The AP metric is reported here. Bolded values show the best results.}
\label{tab:thresh_ablation}
\centering
\small
\setlength{\tabcolsep}{6pt}
\renewcommand{\arraystretch}{1.12}
\begin{tabular}{ccccc}
\toprule
\textbf{$\tau$} & ITODD & IPD & XYZ-IBD & Mean AP \\
\midrule
0.3 & \textbf{46.0} & \textbf{51.5} & \textbf{36.4} & \textbf{44.63} \\
0.4 & \textbf{46.0} & \textbf{51.5} & \textbf{36.4} & \textbf{44.63} \\
0.5 & 45.8 & \textbf{51.5} & 36.3 & 44.53 \\
0.6 & 45.0 & 51.2 & 35.8 & 44.00 \\
\bottomrule
\end{tabular}
\end{table}

\begin{table}
\caption{Ablation study of class-wise NMS threshold. The AP metric is reported here. Bolded values show the best results.}
\label{tab:nms_ablation}
\centering
\small
\setlength{\tabcolsep}{6pt}
\renewcommand{\arraystretch}{1.12}
\begin{tabular}{ccccc}
\toprule
\textbf{$T$} & ITODD & IPD & XYZ-IBD & Mean AP \\
\midrule
0.3 & 44.9 & \textbf{52.3} & 33.9 & 43.70 \\
0.4 & 45.3 & 51.9 & 35.5 & 44.23 \\
0.5 & \textbf{46.0} & 51.5 & \textbf{36.4} & \textbf{44.63} \\
0.6 & 45.9 & 50.8 & 36.3 & 44.33 \\
\bottomrule
\end{tabular}
\end{table}

As shown in Table~\ref{tab:thresh_ablation}, the performance of the proposed pipeline remains stable for lower filtering threshold, with $\tau = 0.3$ and $\tau = 0.4$ resulting in the best performance. 
This threshold controls whether a detection is retained based on its confidence score.
As the threshold increases, the AP score gradually decreases, as valid detections with lower confidence are filtered out.
As shown in Table~\ref{tab:nms_ablation}, the performance of the proposed pipeline shows moderate sensitivity to the NMS threshold, with $0.5$ achieving the highest mean AP of $44.63$. 
Overall, these ablation results demonstrate that the proposed DPM-VFM pipeline is robust to moderate variations in key hyper-parameters, while achieving optimal performance within a consistent operating range.

\subsection{Discussion}
\label{subsec:discussion}

As shown in Table~\ref{tab:bop_industrial_results}, the proposed DPM-VFM framework achieved the highest mean AP score among the compared state of the art methods, demonstrating strong performance on few-shot object detection across the BOP industrial datasets.
The model achieved a particularly high AP score in the IPD and XYZ-IBD datasets, while slightly lower performance was observed in ITODD dataset. 
These results suggest that the proposed method is well-suited for challenging and realistic industrial datasets characterized by clutter, occlusions, and varying lighting conditions.

As in Figure~\ref{fig:qualitative_good_results}, the proposed DPM-VFM framework is able to detect objects accurately and robustly across a variety of industrial scenarios. 
In particular, the method identified a higher number of true positives, including small objects, while producing fewer false positives compared to the state of the art, as in Figure \ref{fig:sota_comparison}. 
This demonstrates the effectiveness of the proposed prototype matching strategy in capturing discriminative object features in difficult environments.
The false positives are primarily caused by errors in object proposal generation, also observed in Figure~\ref{fig:qualitative_bad_results}. 
Specifically, some bounding boxes are either too small or too large, as visible in ITODD samples. 
Furthermore, in some cases, particularly in the IPD dataset, the model detects objects that are not part of the target classes but share similar visual characteristics.
For example, it confuses l-brackets, which is not part of target class, with t-brackets, resulting in false positives. 
The relatively lower AP observed on the XYZ-IBD dataset can also be attributed to these proposal-related issues, where objects are frequently merged with nearby instances or only partially captured, negatively affecting both localization and classification performance.

As shown in Figure \ref{fig:ipd_samples_embeddings}, each object class forms a compact cluster with their embeddings in the feature space. 
In addition to the clear class-wise clustering, we observe the spatial arrangement of the clusters, where the object class with similar geometric properties are clustered nearby in the embedding space.  
For instance, the different classes of corner brackets are located near each other, along with the T-bracket and wraparound bracket. 
Their higher correlation can also be observed in the similarity heatmaps, as shown in Figure~\ref{fig:ipd_prototype_similarity}.
Completely different object classes such as gear and u-bolt are clustered separately.  
This behavior suggests that the pretrained vision foundation model successfully organizes objects based on their geometric and visual similarities.

As shown in Figure~\ref{fig:vfm_ablation}, for all variants of models, larger model variantss generally perform better than smaller models. 
Among all configurations, SAMv1 (ViT-H) with DINOv3 (ViT-L/16) achieves the highest mean AP and is used as the recommended setup. 
Across segmentation backbones, SAMv1 consistently yields the best performance, followed by SAM2.1 and FastSAM. 
In contrast, DINOv3 performs marginally better than DINOv2, suggesting that gains from feature extraction are comparatively limited. 
This implies that the segmentation quality plays a more significant role in overall detection performance.
In all configurations, ITODD and IPD datasets achieve significantly better results than on XYZ-IBD, indicating that dataset complexity and heavy object occlusions remain key challenges despite strong foundation models.

\section{Conclusion}\label{sec:conclusion}
We presented a few-shot object detection framework for industrial applications using vision foundation models.
The proposed DPM-VFM approach combines segmentation-based region proposals with prototype-based similarity matching using feature embeddings from foundation models, enabling object detection with only a few reference sample images.
Experiments on BOP industrial datasets showed that the pre-learned visual representations from the foundation models can be successfully transferred to industrial detection tasks with minimal supervision using the proposed approach.
More importantly, this approach allows for fast on-boarding of new industrial objects with a few example images, without requiring any CAD models or large annotated datasets.
This makes the DPM-VFM framework suitable for real-world industrial use-cases, where new components are frequently introduced and labeled data is limited.

Future work will explore several directions to further improve the proposed framework. 
In particular, extending the method to incorporate depth images or multi-view observations could improve robustness in more complex industrial environments.
Additionally, integrating the proposed detection framework with downstream tasks such as 6D pose estimation or robotic manipulation may enable complete perception pipelines for industrial automation systems.

\bmhead{Supplementary information}

The source code of the proposed solution can be found here: \url{https://github.com/HariPrasanth-SM/DPM-VFM}

\section*{Declarations}

\begin{itemize}
\item Funding: This work was funded by Business Finland under the TwinFlow project (7374/31/2023). 
\item Competing interests: The authors declare that there are no competing interests.
\item Ethics approval and consent to participate: Not applicable. This article does not contain any studies involving human participants or animals performed by the author.
\item Consent for publication: Not applicable.
\item Data availability: The datasets used in this work are publicly available from the Benchmark for 6D Object Pose Estimation (BOP), including the ITODD, IPD, and XYZ-IBD datasets.  
\item Materials availability: Not applicable.
\item Code availability: The source code will be open-sourced upon acceptance, and can be found here: \url{https://github.com/HariPrasanth-SM/DPM-VFM}
\item Author contribution: The authors are responsible and contributed to the conceptualization, methodology, implementation, experiments, and manuscript writing.
\end{itemize}

\clearpage
\bibliography{references}


\end{document}